\documentclass[letterpaper, 10 pt, conference]{ieeeconf}  

\IEEEoverridecommandlockouts                              

\usepackage{graphics} 
\usepackage{epsfig} 
\usepackage{mathptmx} 
\usepackage{times} 
\usepackage{amsmath} 
\usepackage{amssymb}  
\usepackage{color}
\usepackage{tabularx}
\usepackage[hyphens]{url}
\usepackage[hidelinks]{hyperref}
\hypersetup{breaklinks=true}

\title{\LARGE \bf
Variable-Scaling Rate Control for Collision-Free Teleoperation of an  Unmanned Aerial Vehicle
}

\author{Dawei Zhang$^{1}$ and Rebecca P. Khurshid$^{2}$
\thanks{$^{1}$Dawei Zhang is with the Department of Mechanical Engineering, Boston University, Boston, MA 02215, USA
        {\tt\small dwzhang@bu.edu}}%
\thanks{$^{2}$Rebecca P. Khurshid is with the Department of Mechanical Engineering and the Division of Systems Engineering, Boston University, Boston, MA 02215, USA
        {\tt\small khurshid@bu.edu}}%
}

\begin{document}

\maketitle
\thispagestyle{empty}

\pagestyle{empty}

\begin{abstract}
We propose that automatically adjusting the scale factor in rate-control teleoperation could enable a human operator to better control the motion of a remote robot. In this paper, we present four new variable-scaling rate-control methods that adjust the scale factor depending on the state of the user's input commands and/or the risk of a collision between the robot and its environment. Methods that depend on the risk of a collision are designed to guarantee collision avoidance by setting the scale factor to be zero if the operator issues a command that would result in a collision between the robot and its environment. A within-subject user study was conducted to determine the effects 
 of the four newly designed rate-control methods and a traditional fixed-scale rate-control method on a person's ability to complete a navigation task in a simulated two-dimensional environment. The results of this study indicate that well-designed variable-scale rate control can guarantee collision-free teleoperation without reducing task efficiency. 

\end{abstract}

\section{INTRODUCTION}

A small and agile unmanned aerial vehicle (UAV) can be used by rescue teams to search for survivors in the wake of a man-made or natural disaster. For example, a rescue worker could fly the UAV through a building that is unsafe or impossible for the rescue team to enter directly themselves. The rescue worker should be able to quickly fly the UAV, while deftly maneuvering around objects in its environment. The rescue worker must also be able to precisely control the UAV's motion if he or she wants to carefully inspect a certain area. 




To control the motion of the remote robot, the human operator issues commands using a control interface, such as a joystick. To enable the operator to control the motion of the robot over large distances using a much smaller control interface, it is necessary to modify the operator's input commands through a forward-control method to calculate the desired state of the robot. Rate control, also known as velocity control, is the most common control method used to remotely control unmanned ground or aerial vehicles \cite{telerobotics}. Under this method, the robot's commanded velocity is proportional to the position of the control interface. While rate control enables an operator to span large areas with the remote robot, a  major limitation is that it can be difficult for the operator to precisely control the position of the remote robot through rate control teleoperation~\cite{spanning,mappings}. If the remote robot can move quickly, as is the case for agile UAVs, the operator could easily crash the remote robot. Thus, although the methods developed in the paper could be applied to any remote mobile robot, our intended application is the remote control of agile UAVs. 

Several researchers have implemented haptic feedback schemes as a means to reduce collisions between a remotely controlled UAV and its environment \cite{interface,hapticsUAV,bilateral,bilateral2}. Under these schemes, a grounded kinesthetic (force-reflecting) control interface is used to apply a force on the user when there is an increased risk of a collision. Typically, the magnitude of the force is related to the risk of a collision and the direction of the force is pointed directly away from the object that poses the greatest risk. For example, Brant and Colton set the magnitude of the force of the haptic feedback to be proportional to the time that it would take the UAV to collide with an object in its environment, if the UAV continued flying with its current velocity \cite{hapticsUAV}. The results of a user study showed that this method was effective in reducing the number of collisions between the robot and its environment, without sacrificing task efficiency \cite{hapticsUAV}. Drawing on potential functions used in robotic path planning, Lam et al. proposed a parametric risk field to calculate the risk of a collision, which was then used to generate the magnitude of the force exerted on the user through the control interface \cite{Lam2009}. Hou and Mahony implemented a similar method using an admittance-type haptic device to physically prevent the operator from issuing a command that would result in a collision \cite{DKBC}.

Alternatively, changing the mapping between the operator's input commands and the commanded state of the remote robot can help improve the operator's ability to control the motion of a remote robot. Hybrid forward-control schemes have been implemented to automatically switch between enabling the operator to directly control the robot's velocity, which is better for large movements, or enabling the operator to directly control the robot's position, which enables more precise control of the robot's position \cite{mappings,switching}. Romano et al. proposed a hybrid control law that commands the robot's velocity to be proportional to the square of the velocity of the control interface, so that slow movements of the control interface enable the operator to precisely control the position of the robot and fast movements of the control interface enable the operator to move the robot quickly \cite{needles}. 

In this paper, we present novel variable-scaling rate control methods to enable both fine and fast control of a robot by automatically changing the scale factor relating the position of the control interface to the commanded velocity of the remote robot. The scale factor is adjusted based on the human operator's input commands to allow for finer motion control when the operator issues smaller input commands. The scale factor is also adjusted based on the risk of the UAV's collision in a manner that guarantees that the robot cannot be commanded to collide with an object in its environment. 

We note that shared-autonomy methods also use information related to the user's input and the robot's environment to help an operator control the motion of a remote robot (or virtual agent). A leading shared autonomy paradigm involves the robot predicting the operators goal (or a probability distribution of the operator's goal) online and acting semi-autonomously to help achieve the predicted goal \cite{dragan2012formalizing,hauser2013recognition,javdani2015shared,muelling2015autonomy}. Another method helps ensure safety even when the operator commands an unsafe action, by having the robot move to the closest state to the human's command that satisfies some safety criteria \cite{schwarting2017parallel}. Another method uses a policy trained via deep reinforcement learning to alter the operator's commands, if they are deemed to be sufficiently suboptimal, to a sufficiently near-optimal action closest to the user's input \cite{reddy2018shared}. In these state-of-the-art shared autonomy systems, the human operator's input is used to generate intermediate commands, which are then blended with or replaced by the commands generated by the shared-autonomy method. In this paper, we seek to improve a direct mapping between the user's input and the commands sent to the robot. 

A detailed description of our methods is presented in Section \ref{sec:methods}. Section \ref{sec:user study} describes the design and results of a user study investigating the effects of variable-scaling rate control on an operator's ability to control a simulated robot in a two-dimensional environment. We interpret the results of this user study in Section \ref{sec:discussion}. Finally, Section \ref{sec:conclusion} presents the main conclusions of this paper and our plans for future research.

\section{VARIABLE-SCALING RATE CONTROL} \label{sec:methods}
Variable-scaling rate-control builds on the classical rate-control method. Rate control in robotic teleoperation is used to map the position of the operator's control interface to the desired velocity of the remote robot, which can be described as follows:

\begin{equation} \label{eq:standardRate}
\vec{V}_{d,R} = S*\vec{P}_{i}.
\end{equation}
In this equation the commanded velocity of the robot, $\vec{V}_{d,R}$, is proportional to the position of the control interface, $\vec{P}_{i}$, through a proportionality constant, $S$. Picking an appropriate scale factor, $S$, can be challenging. If the scale factor is too small, the motion of the robot will be slow and the operator will need to spend more time and energy to move the robot to the goal position, which may be frustrating. On the other hand, if the scaling is too large, it will be hard for the operator to precisely control the position of the robot. A large scale factor can also increase the likelihood that an operator would crash the remote robot, especially if the remote robot can move quickly, as is the case for agile UAVs.

In this paper, we automatically adjust the value of $S$ based on the user's input and the risk of a collision between the remote robot and its environment. Namely, we multiply a constant scale factor, $S_c$, by a scale factor related to the human's input, $S_{human}$, and a scale factor relating to the risk of a collision, $S_{risk}$:
\begin{equation} 
S = S_c*S_{human}*S_{risk}
\end{equation}
\subsection{User's Input}
The scale factor related to the human's input should allow for fast control when the human commands a large velocity and should allow for fine control when the human commands a smaller velocity. In this implementation, we use $S_{human}$ to reduce the robot's commanded velocity if the user is displacing the control interface less than some distance, $P_c$, indicating that the human is trying to precisely control the position of the robot. If the human displaces the the control interface greater than $P_c$, then $S_{human}$ is equal to 1. The scale factor related to the human's input is represented by:
\begin{equation} 
S_{human} = \min(1, \frac{\left | \vec{P}_{i} \right |}{P_c} ) 
\end{equation}

\subsection{Risk of Collision}
The scale factor relating to the risk of a collision, $S_{risk}$, decreases the commanded velocity as the risk that the UAV will collide with another object in its environment increases. This can be described as:

\begin{equation}
 S_{risk} = 1-C_r,   
\end{equation}

where $C_r$ represents the likelihood, between 0 and 1 inclusive, that the human will command the robot to collide with another object in its environment. If the risk of a collision is 0, then $S_{risk}$ will be equal to 1 and the commanded speed of the robot will not be reduced. If the robot is certain to collide with another object, then the risk of a collision will be equal to 1 and $S_{risk}$ will be equal to 0. Thus, the robot's commanded velocity will be equal to 0, preventing a collision from occurring. 

\begin{figure}[t] 
      \centering

      \includegraphics[width=0.85\columnwidth, trim={0cm 0.8cm 0 1cm},clip]{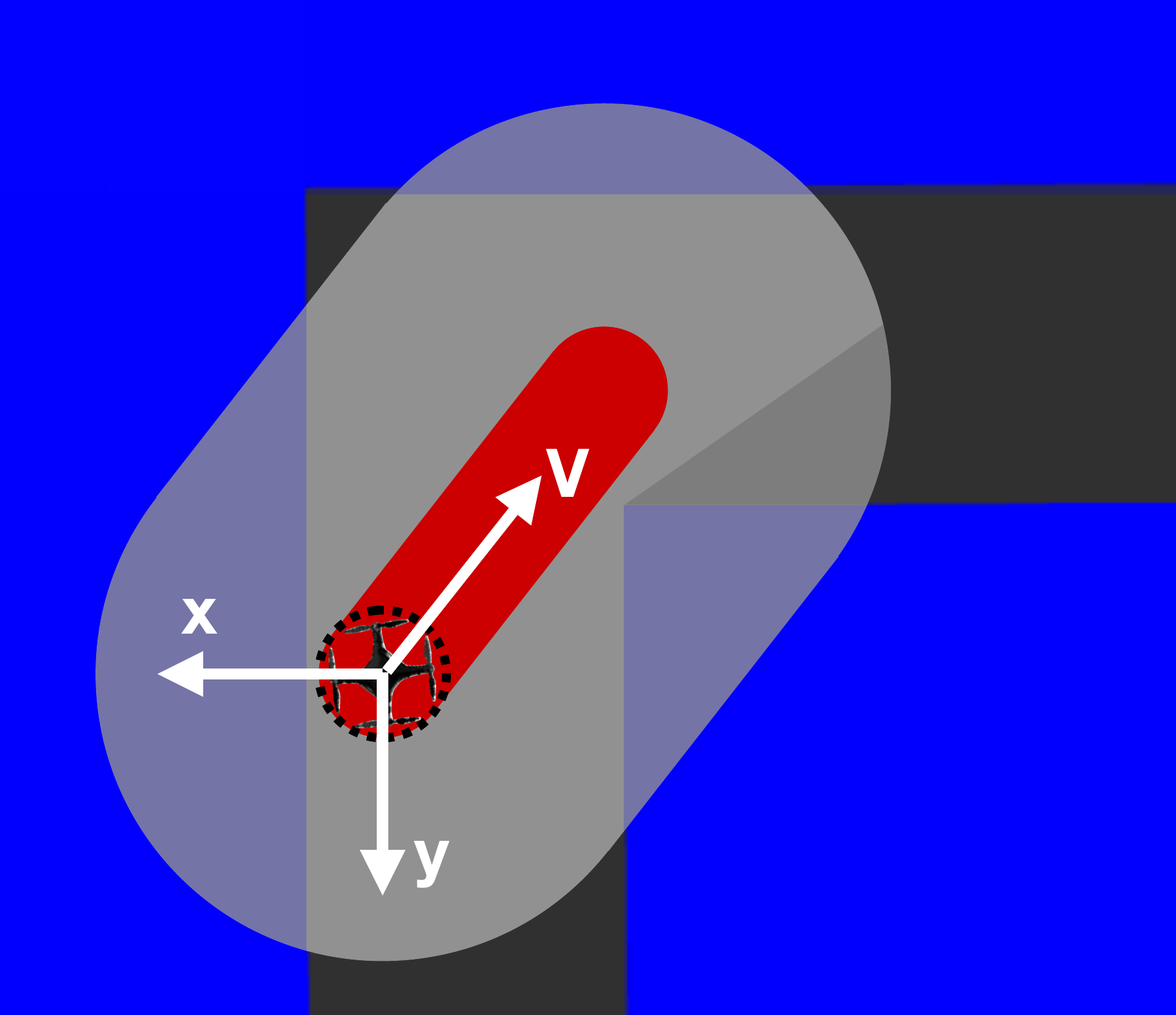}
      \caption{\small A UAV flown with velocity, V, through a hallway (shown in black) would unavoidably collide with any object located in the red region and may collide with any object located in the gray region. Transparent gray regions represent areas where a line-of-sight sensor mounted on the UAV would not be able to gather data.}
      \label{fig:potential}
      \vspace{-15pt}
   \end{figure}

We adopt the parametric risk field, developed by Lam et al. \cite{Lam2009}, to calculate  the risk factor, $C_r$. Following, \cite{Lam2009}, we first calculate a critical region, in which a collision will be unavoidable, represented by the red region in Fig \ref{fig:potential}. The critical region depends on the robot's current velocity and maximum acceleration. For a UAV, the critical region includes the space directly around the UAV, which is circumscribed by a circle that has a radius, $R_{UAV}$, which is shown by the dashed black circle in Fig \ref{fig:potential}. The critical region also includes the space swept out by this circumscribing circle if the UAV were to decelerate as quickly as possible. The length of the critical region can be calculated by: 
\begin{equation}
L_{cr}=2R_{UAV}+\frac{\left | \vec{V} \right |^{^{2}}}{2a_{max}}   
\end{equation}
where $\vec{V}$ is the UAV's current velocity and $a_{max}$ is the magnitude of UAV's maximum acceleration. 

If an object is located just outside the critical region, there is a high risk of collision between the UAV and that object. If an object is located far away from the critical region, there is a low risk of collision between the UAV and the object. Thus, Lam et al. proposed determining the risk of a collision using a potential field in the space around the critical region \cite{Lam2009}. In this implementation, we have chosen to compute the potential field at all points within a distance, $d$, from the critical region. Because objects pose a lower risk at low velocities and a higher risk at higher velocities, we compute $d$ as follows:
\begin{equation}
d = d_c + s_d\left | \vec{V} \right |
\end{equation}
where $d_c$ and $s_d$ are constant values. The region over which the potential field will be computed is shown by the transparent gray region in Fig \ref{fig:potential}.

The risk of a collision varies from 0, at the far extent of the potential field, to 1, at the boundary between the potential field and the critical region. The risk of collision between the UAV and a point on a obstacle, $p$, that is some distance, $d_o$, outside the critical region, can be computed by:
 
\begin{equation} \label{eq:crp}
C_{r,p}=\begin{cases}
0, &d_o > d \ \\
 1,&d_o < 0 \\ 
f(d,d_o,V, a_{max})\ |\  f \in [0,1] & otherwise
\end{cases} 
\end{equation} 

The function $f$ can be any smooth function that ranges from 0, when $d_o = d$ to, 1, when at the border of the critical region. In this implementation, $f$ can be expressed by $\frac{d-d_0}{d}$. In this formulation, the point of occupied space that is closest to the critical region poses the highest risk of a collision. To calculate the overall risk of a collision of the UAV, one option is to set $C_{r}$ to be that maximum value of $C_{r,p}$ over all occupied points in the environment, such that:

\begin{equation} \label{eq:maxcrp}
C_r = \max\limits_{p} C_{r,p} 
\end{equation} 

Note, that because the risk of a collision will only be equal to 1 when an object is located at the boundary of the critical region, this method will allow the UAV to get arbitrarily close of objects in its environment, although will force the UAV to approach these objects at slow speeds. 

Theoretically, the UAV could become stuck if an object is located exactly at the boundary of the critical region, so that the commanded velocity would become zero and the operator would not be able to move away from the object. This is not likely a concern in practice because the commanded velocity of the robot is very small when obstacles are near the boundary of the critical region, so it takes considerable effort to force an object to be just at the border of the critical region. However, a practical concern of this implementation is that when the overall risk of a collision becomes high, the commanded velocity of the robot will be small, even when the operator is trying to move the robot away from the object. We address this limitation by introducing direction-dependent scaling methods to calculate $S_{risk}$.

\subsection{Direction-Dependent Scaling} 
The risk of a collision between the UAV and it's environment is dependent both on the location of objects around the UAV and the direction of the UAV's commanded velocity. For example, consider the scenario shown in Fig. \ref{fig:potential}, in which the operator is flying the UAV in a hallway. In this scenario, the individual point with the highest risk factor is located on the wall closest to the UAV. However, the operator is commanding the robot to fly away from this direction, and thus the actual risk of a collision is lower than the maximum value of $C_{r,p}$. In this example, it it clear that it could be beneficially to independently change the mapping scale in the X and Y directions. If the scale factor is independently varied in the X and Y directions, then fast control can occur along the direction of the hallway while fine control occurs in the direction of the nearest obstacle, i.e. the wall. 

At each time step, we define the X-axis to be in the direction of the point that has the maximum risk, $C_{r,p}$, as determined by Equation \ref{eq:crp}. The Y-axis is set to be perpendicular to the X-axis, such that the direction of the commanded velocity, the X-axis, and the Y-axis are all coplanar. We can then extend Equation \ref{eq:standardRate} to: 

\begin{equation} \label{eq:directionalScaling}
\begin{bmatrix}
V_{R,x} 
\\ 
V_{R,y} 
\end{bmatrix} = \begin{bmatrix}
S_x*P_{i,x}

\\ 
S_y*P_{i,y}
\end{bmatrix} 
\end{equation}
where $S_x$ and $S_y$ are computed as: 
\begin{equation} 
S_x = S_c*S_{human}*S_{risk,x}
\end{equation}
\begin{equation} 
S_y = S_c*S_{human}*S_{risk,y}.
\end{equation}

In the above equations, all components are with respect to the local X-Y frame, whose X-axis points in the direction of the occupied point with the highest $C_{r,p}$. We note that no scale factor is needed in the Z-axis, because there is no component of the commanded velocity in the Z-axis. 

$S_{risk,x}$ and $S_{risk,y}$ are calculated as $(1-C_{r,x})$ and $(1-C_{r,y})$, respectively. Similar to the overall risk factor, $C_{r,x}$ is calculated as: 

\begin{equation} \label{eq:maxcrpx}
C_{rx} = \max\limits_{p_x} C_{r,p,x}, 
\end{equation} 
where $p_x$ is the set of points such that a line through the point in the X-direction would intersect with the critical region. $C_{r,p,x}$ is calculated by:

\begin{equation} \label{eq:crpx}
C_{r,p,x}=\begin{cases}
0, &d_{o,x} > d \ \\
 1,&d_{o,x} < 0 \\ 
f(d,d_{o,x},V, a_{max})\ |\  f \in [0,1] & otherwise.
\end{cases} 
\end{equation} 
Here, $d_{o,x}$ is the magnitude of the x-component of the vector between point $p$ and the point on the critical region closest to $p$. 

Similarly, $C_{r,y}$ is calculated as: 

\begin{equation} \label{eq:maxcrpy}
C_{r,y} = \max\limits_{p_y} C_{r,p,y}, 
\end{equation} 
where $p_y$ is the set of points such that a line through the point in the Y-direction would intersect with the critical region. $C_{r,p,y}$ is calculated by:

\begin{equation} \label{eq:crpy}
C_{r,p,y}=\begin{cases}
0, &d_{o,y} > d \ \\
 1,&d_{o,y} < 0 \\ 
f(d,d_{o,y},V, a_{max})\ |\  f \in [0,1] & otherwise.
\end{cases} 
\end{equation} 
Here, $d_{o,y}$ is the magnitude of the x-component of the vector between point $p$ and the point on the critical region closest to $p$. In Equations \ref{eq:crpx} and \ref{eq:crpy} the function $f$ is $\frac{d-d_{o,x}}{d}$ and $\frac{d-d_{o,y}}{d}$, respectively.


The direction of the commanded velocity, $\vec{V}$, can also be taken into account when calculating $C_{r,x}$ and $C_{r,y}$. Again, referring to Fig. \ref{fig:potential}, we see that the X-component of the commanded velocity is in the opposite direction of the point with the highest overall $C_{r,p,x}$. Therefore, is may make sense to only calculate $C_{r,x}$ using objects only in the negative X direction from the critical region. We can formalize this as:


 \begin{equation} \label{eq:r3x}
C_{r,x}=\begin{cases}
\max\limits_{p_{x+}} C_{r,p,x} , &\vec{V}\cdot \hat{x} > 0\ \\
\max\limits_{p_{x-}} C_{r,p,x},& \vec{V}\cdot \hat{x} < 0,
\end{cases}
\end{equation}
where $p_{x+}$ and $p_{x-}$ are the subsets $p_{x}$ of points with X position coordinates in the positive and negative X directions, respectively, and $\hat{x}$ is the unit vector in the X direction. Similarly, we have that
 \begin{equation} \label{eq:r3y}
C_{r,y}=\begin{cases}
\max\limits_{p_{y+}} C_{r,p,y} , &\vec{V}\cdot \hat{y} > 0\ \\
\max\limits_{p_{y-}} C_{r,p,y},& \vec{V}\cdot \hat{y} < 0.
\end{cases}
\end{equation}

\section{USER STUDY} \label{sec:user study}
We conducted a user study to understand the effects of variable-scaling rate control on a user's ability to control a UAV. This study was approved as exempt by the Boston University Institutional Review Board under protocol number 5070E. 

\subsection{Experimental Setup}
\begin{figure}[t] 
      \centering

      \includegraphics[width=\columnwidth, trim={0cm 0.8cm 0 1cm},clip]{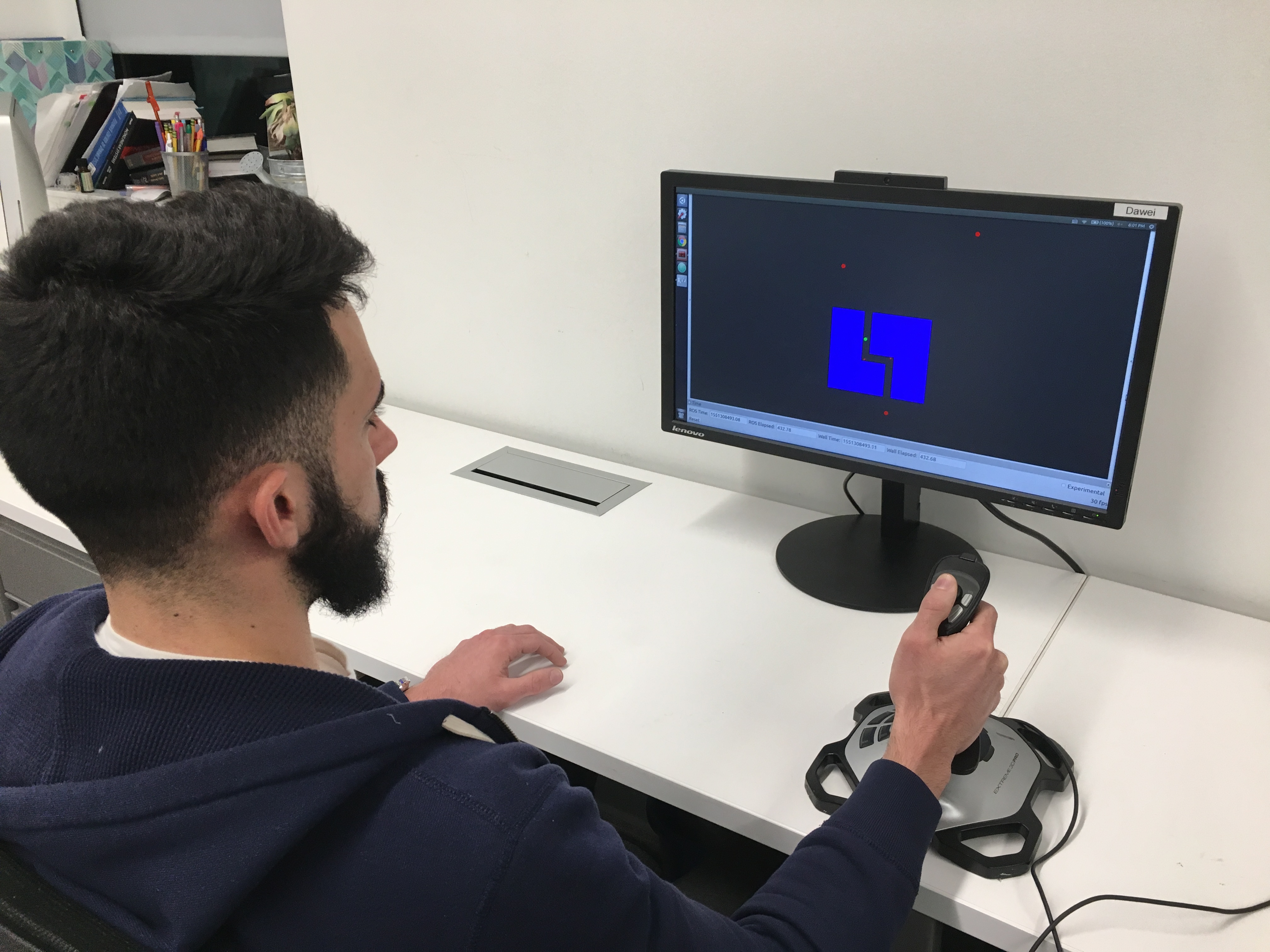}
      \caption{\small  Each participant used a joystick with their right hand to control the motion of a simulated robot in a 2D environment.}
      \label{fig:Joystick}
      \vspace{-5pt}
   \end{figure}

As shown in Fig. \ref{fig:Joystick}, each subject controlled the motion of a virtual robot in a simulated 2D environment. The control interface was a Logitech Extreme 3D Pro joystick. The position of the joystick is measured in terms of its maximum displacement, so that a reading of 0 corresponds to the neutral position and a reading of 1 corresponds to maximum displacement. Robot Operating System (ROS) \cite{ros} was used to create a 2D simulated environment and a simulated point robot. The radius of the point robot is $0.2$ m and it's maximum acceleration is $35$ m/s$^2$, based reported values for a high-power quadrotor UAV \cite{UAVacceleration}. All the values of parameters used in the user study are listed in Table. \ref{tab:parameters}. 

\begin{figure}[t] 
      \centering

      \includegraphics[width=\columnwidth, trim={0cm 0cm 0 0cm},clip]{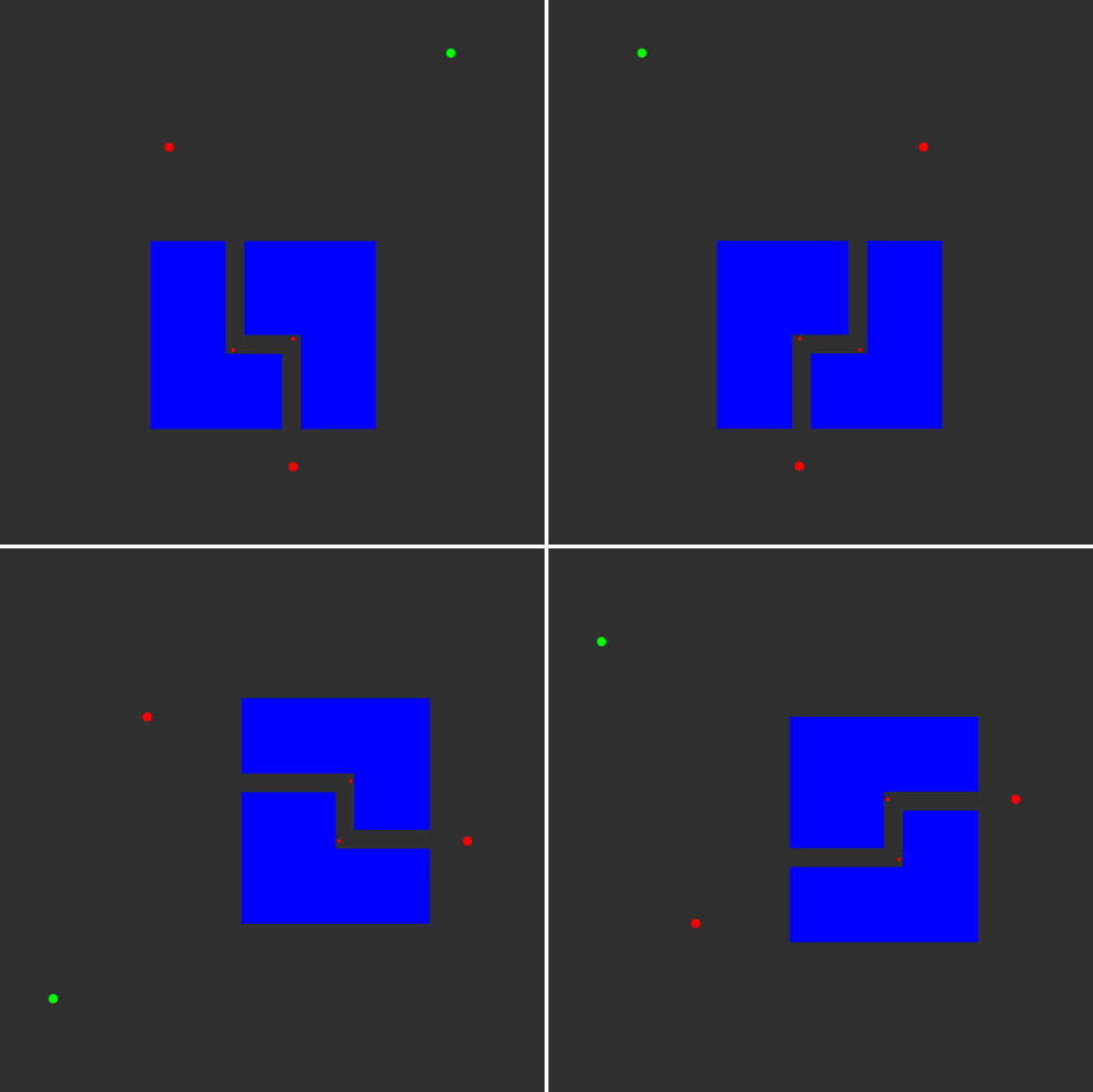}
      \caption{\small The four simulated 2D environments used in this study. The simulated robot was shown to the users as a green circle. In the above figures, the robot is at its starting location. During each trial, the subject uses a joystick to move the simulated robot to a target area in free space, then to two target locations in the simulated hallway, then to a final goal location near the exit of the hallway. Target locations were shown to the user as red circles.}
      \label{fig:environment}
      \vspace{-15pt}
   \end{figure}

Four simulated environments, shown in Fig. \ref{fig:environment}, were used in this study. Each environment contains open space near the starting location of the simulated robot, a constrained hallway with two turns, and free space on the other side of the hallway. The width of the hallway is $1$ m. There is one goal location in the free space close to the starting location, two target locations in the corners of the hallway, and a final target position at the end of the hallway. Participants had a full overhead view of the simulated environment. The simulated walls did not constrain the robot's position, meaning that the person could command the position of the robot to penetrate the virtual wall. 

The velocity of the simulated robot was set to be the desired velocity determined by the forward-control method, unless doing so required the simulated robot to accelerate faster than its maximum acceleration. In that case, the acceleration of the simulated robot was set to be the maximum acceleration in the direction needed to achieve the desired velocity. The position of the simulated robot was updated at a rate of 100Hz through Euler integration. 

\setlength{\extrarowheight}{1pt}
\begin{table}
    \centering
     \caption{Parameters in User Study}
    \label{tab:parameters}
    \begin{tabular}{ |c|c|c|c|c|c|c|} 
  \hline
  
 Parameter & $S_c$ & $P_c$& $R_{UAV}$&$a_{max}$&$d_c$&$s_d$ \\ 
  \hline

 Value &5.0&0.5&0.2&35&0.3& 1.0\\ 
 \hline
 Units &-&-&$m$&$m/s^2$&$m$&-\\
 \hline
\end{tabular}
\end{table}

\subsection{Evaluated Rate-Control Methods}
Each subject tested the following five rate-control methods:
\begin{itemize}
\item \textbf{(C)} Constant scale factor: 
\begin{equation}
S_x = S_y = S_c,
\end{equation} 
where $S_x$, $S_y$ refer to the scale factors in Equation \ref{eq:directionalScaling}.
\item \textbf{(H)} Scale factor based on the user's input: 
\begin{equation}
S_x = S_y = S_c*S_{human}
\end{equation}
\item \textbf{(R1)} Scale factor based on the user's input and the risk of collision:
\begin{equation}
S_x = S_y = S_c*S_{human}*S_{risk},
\end{equation}
where $S_{risk}$ is calculated from Equation \ref{eq:maxcrp}.
\item \textbf{(R2)} Scale factor based on the user's input and the risk of collision in the X- and Y- directions, separately:
\begin{equation}
S_x = 1-C_{r,x} 
\end{equation}
\begin{equation}
S_y = 1-C_{r,y},
\end{equation}
where $C_{r,x}$ is determined by Equation \ref{eq:maxcrpx} and $C_{r,y}$ is determined by Equation \ref{eq:maxcrpy}.

\item \textbf{(R3)} Scale factor based on the user's input and the risk of collision in the X- and Y- directions, separately, accounting only for the risk of collision with objects in the direction of the commanded velocity:
\begin{equation}
S_x = 1-C_{r,x} 
\end{equation}
\begin{equation}
S_y = 1-C_{r,y},
\end{equation}
where $C_{r,x}$ is determined by Equation \ref{eq:r3x} and $C_{r,y}$ is determined by Equation \ref{eq:r3y}.

\end{itemize}


\subsection{Subject Population}
Fifteen subjects participated in this user study (three female and twelve male). All of the subjects were right-handed and between the ages of 21 and 27.

\subsection{Experimental Procedure}
A within-subject experimental protocol was used. Each subject completed 8 trials with each of the five rate-control methods. For each rate control method, the subject navigated the simulated robot through each of the four simulated environments two times. The presentation order of the rate-control methods was counterbalanced using a Latin Square. The presentation order of the environments was randomized. Participants used the joystick with their right hand to navigate the simulated robot to each of the four target locations. Subjects pressed a button on the control interface to indicate when they felt they had reached each target location. The trial began when the participant issued the first command to the simulated robot. The trial ended when the participant pushed the button at the final target location. Participants were told to complete the task as quickly as possible, without colliding the robot with the simulated walls in the experiment. 

After completing 8 trials for each method, subjects provided subjective measures of their experience using NASA Task Load Index (NASA-TLX) \cite{hart1988}. After all five forward-control methods were tested, subjects completed a final survey ranking each of the methods according to: 
\begin{itemize}
\item their favorite method
\item the method they would choose to do a more complicated task
\item the method they would choose to accomplish a task quickly.
\end{itemize}

Methods were referred to by their presentation order. No information about the control methods was given, beyond the fact that the position of the joystick would be mapped to the robot's velocity. 

          \begin{figure*}[t] 
      \centering

      \includegraphics[width=7in,trim={0cm 2.6cm 0 0cm},clip]{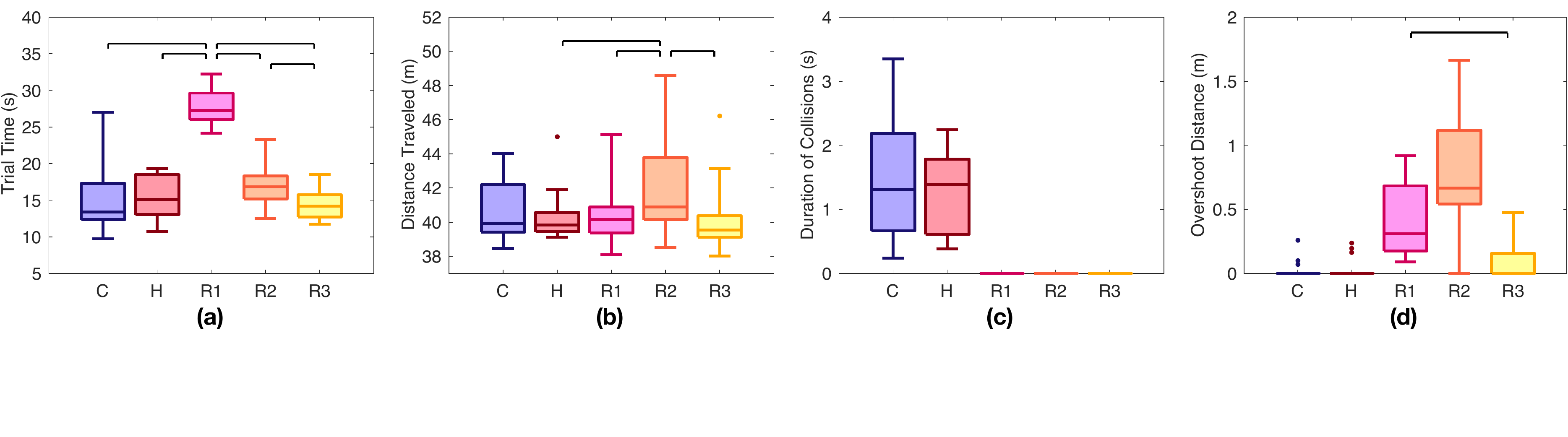}
      \vspace{-15pt}
      \caption{\small Results of the user study as measured by (a) trial time, $T_{trial}$,  (b) total distance traveled by the robot, $D_{total}$. (c) duration of collision, $T_{collision}$, and (d) the overshoot distance of the final target, $D_{overshoot}$. Significant pairwise differences between the different rate-control methods are marked with brackets. For all metrics, lower values indicate better performance.}
      \label{fig:metricResults}
   \end{figure*}
   
             \begin{figure*}[t] 
      \centering

      \includegraphics[width=7in,clip]{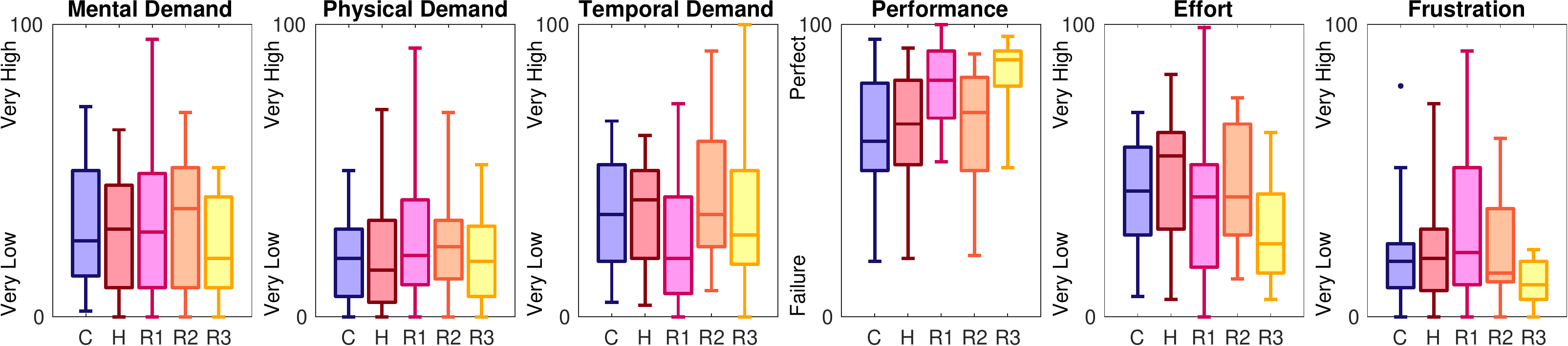}
      \caption{\small  Subjective workload and performance ratings as measured via the NASA-TLX survey.}
      \label{fig:resultsTLX}
      \vspace{-15pt}
   \end{figure*}
   
\subsection{Data Analysis}
During each trial we recorded the simulated robot's position at a rate of 100Hz. We use the following metrics to evaluate the user's performance when using each of the five rate-control methods:
\begin{itemize}
\item $T_{trial}$: Trial duration, which is the time in seconds that each subject spent to complete the task. A smaller $T_{trial}$ implies a better performance in speed.
\item $D_{total}$: Total distance traveled by the virtual UAV during the trial. A small value of $D_{total}$ means better economy of motion.
\item $T_{collision}$: The total duration of time that the simulated robot was in contact with the simulated walls. A small value of $T_{collision}$ is preferable.
\item $D_{overshoot}$: The distance that the robot traveled past the final target, after it left the simulated hallway. A small value of $D_{overshoot}$ indicates better performance.
\end{itemize}

These four metrics were averaged over the eight trials conducted by each subject for each rate-control method. Repeated measures analysis of variance (rANOVA) was used to determine whether the rate-control method had any effect on task performance. When a significant difference in subject performance was found, Tukey's test was performed at a confidence level of $\alpha = 0.05$ to determine which methods led to significant differences in the metric. All data analysis was performed using MATLAB's built-in statistical functions.


\subsection{Experimental Results}
The quantitative results of the user study are shown in Fig \ref{fig:metricResults}. In these plots, significant pairwise differences are marked with brackets.  The rate-control method used has a significant effect on the trial time (F(4,56)=99.82, p $<$ 0.0001). Participants took significantly longer to complete the task under Method R1, which is the most conservative method, as compared to the other four methods. It also took participants significantly longer to complete the task using Method R2, which sets different scale factors in the X- and Y- directions, than when using Method R1, which accounts for the direction of the desired velocity when setting the X- and Y- scale factors. 

 The rate-control method used has a significant effect on the total distance traveled by the simulated robot (F(4,56)=8.52, p $<$ 0.0001). The distance traveled by the robot when under Method R2, was significantly longer than the distance traveled under Methods H, R1, and R3. There is no significant differences in $D_{total}$ when comparing Methods C,H,R1 and R3 against each other. 
 
 There were no collisions between the robot and its environment when the participants used Methods R1, R2, and R3 because these methods guarantee collision-free teleoperation.  For the metric of $T_{collision}$ a Student's T-Test was performed to evaluate any difference between method C and H, because collision duration as exactly equal to zero for methods R1, R2, and R3. No significant difference was found between Methods C and H, in terms of collision duration. 
 
 As shown in Fig. \ref{fig:metricResults}(d), only three operators overshot the target just after the exit of the hallway when using Methods C and H. Because the overshoot for Methods C and H were heavily saturated at zero, only differences in R1, R2, and R3 were analyzed. The results of the rANOVA run on these three methods show that the rate-control method did have an effect of overshoot (F(2,28) =  26.183, p $<$ 0.0001). Participants were less prone to overshoot the final target when using Method R3, as compared to Method R1.

\begin{figure}[t] 
      \centering
      \includegraphics[width=\columnwidth, clip]{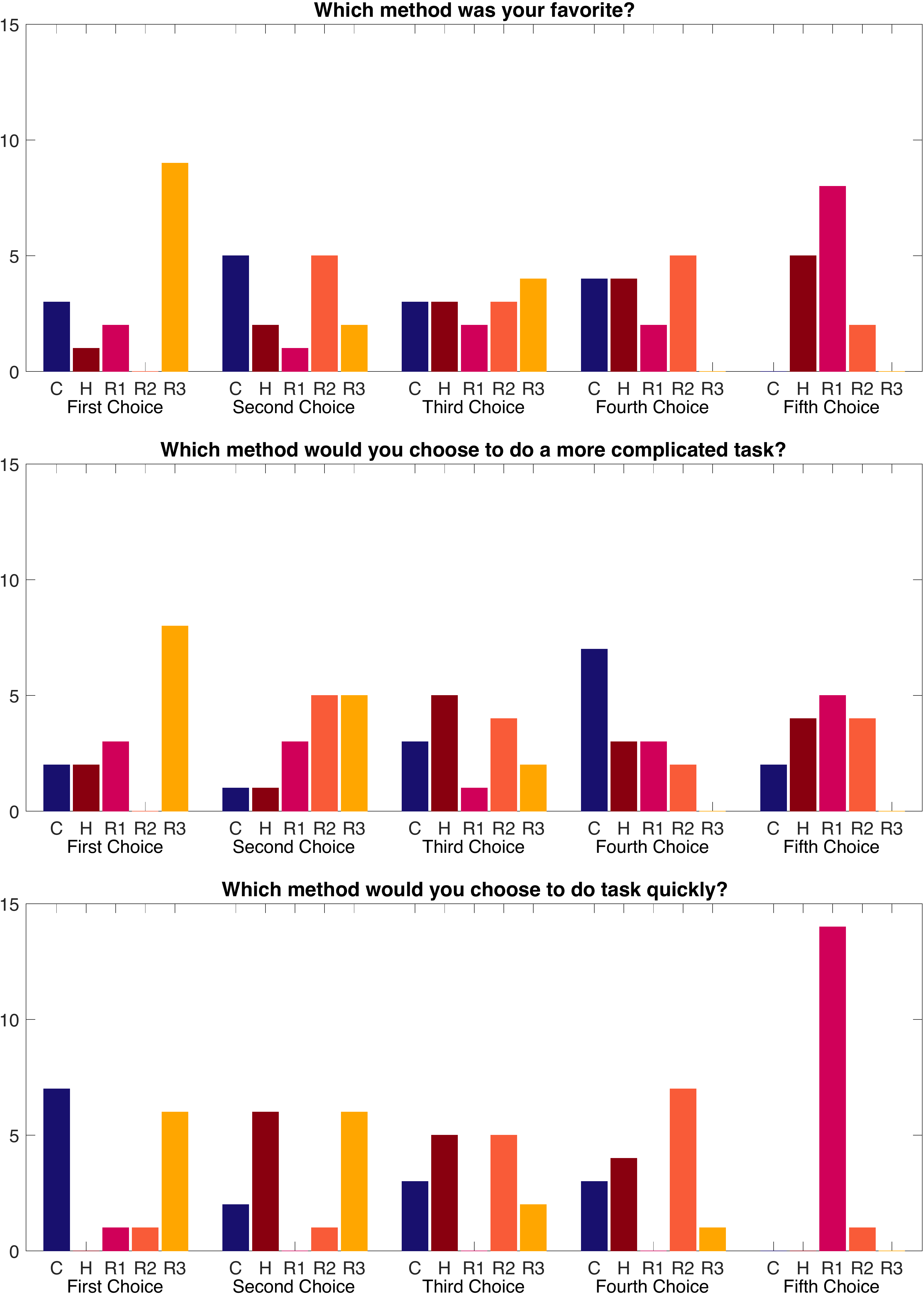}
      \caption{\small User rankings of the five rate-control methods according to their favorite method (top), the method they would choose to do a more complicated task (middle), and the method they would choose to do a task quickly (bottom). The bars represent the number of participants who chose each method as their first choice, second choice, third choice, fourth choice, and fifth choice.}
      \label{fig:surveyData}
      \vspace{-15pt}
   \end{figure}

The different rate-control methods did not result in a difference in the subjective rating of workload and task performance, as measured by subject responses to the NASA-TLX, which are shown in Fig. \ref{fig:resultsTLX}. The subjects' rankings of the five rate-control methods are shown in Fig. \ref{fig:surveyData}. Nine of the fifteen study participants chose Method R3 as their favorite method and seven participants indicated that they would choose Method R3 to do a more complicated task. No one ranked Method R3 as their least or second to least favorite method. Furthermore, no one ranked Method R3 as fourth or fifth choice to complete a more complicated task. Method C and Method R3 were the top two choices for the preferred method to complete a task quickly. Fourteen of fifteen users indicated that Method R1 would be their last choice to complete a task quickly. While many subjects ranked Method R1 as their least favorite, two participants indicated that it was their favorite.

\section{DISCUSSION} \label{sec:discussion}

The results of this study indicate that variable-scaling rate control can improve a person's ability to control a UAV. Importantly, we note that by taking into account the risk that the robot will collide with its environment, variable-scaling rate control can guarantee that the human operator will not crash the remote robot. Calculating a single scale factor based on the overall highest risk of a collision, as is done in Method R1, reduces the speed at which the operator can fly the UAV in constrained areas. This is reflected by the increased time it took the operators to complete the task when using Method R1. Calculating separate scale factors for the direction pointing to the object in the UAV's environment posing the biggest collision risk and a direction perpendicular to this, as is done in R2, can significantly increase the speed with which the user can fly the UAV. This is especially true for a hallway scenario, where the distance between the UAV and a wall will typically be much smaller than the distance between the UAV an object along the length of the hallway. Moreover, calculating separate scale factors for two perpendicular directions accounting only for the risk of collision with objects in the direction of the commanded velocity, as is done in Method R3, resulted in faster task completion times than considering the overall all risk in these directions. Notably, the results of the user study indicated that Method R3 did not effect the user's task completion time, as compared to Methods C and H, which did not slow the UAV down when the risk of a collision was high. The finding that Method R3 did not result in decrease task efficiency is also reflected in the fact that nearly as many study participants indicated that they most prefer to complete a task quickly using this method, as compared to the number of participants who selected the method that never reduced the scale factor. 

One limitation of reducing the scale factor when the risk of collision is high, is that our results indicate that it may be difficult for people to control the position of the robot when the risk of collision transitions from high to low. In this study, under Methods R1, R2, and R3, when the UAV exited the hallway, the same position of the joystick will suddenly result in a higher commanded speed to the robot. Nearly all participants overshot the target located near the hallway's exit under Methods R1 and R2, and about half of the participants overshot this target when using Method R3. Only three participants overshot this target when using the conditions that did not reduce the scale factor based on the risk of a collision between the UAV and an object in its environment. This indicates that there should be limits on the rate with which the overall rate-control scale factor can be increased. 

There are no differences between user performance when using the rate-control method that reduced the scale factor based on the user's input, H, and the user performance when using the rate-control method that never reduced the scale factor, C. However, participants were still likely to be able to perceive a difference between these conditions, as indicated by the fact that seven of the fifteen participants indicated that they would most prefer to use Method C to complete a task quickly, while no participants indicated that they would most prefer to used Method H.

Based on the results of the user study, Method R3, which adjusts the rate-control's scale factor based on the user's input and the risk of collision in the X- and Y- direction accounting only for the risk of collision with objects in the direction of the commanded velocity, has a best overall performance.  
\addtolength{\textheight}{-4.5cm}

\section{CONCLUSIONS AND FUTURE WORK} \label{sec:conclusion}

In this paper, we tested the hypothesis that automatically adjusting the scale factor in rate-control teleoperation would better enable the operator to control the motion of a robot. We developed methods that reduce the scale factor when the user is issuing small velocity commands and the risk of a collision between the robot and objects in its environment is high. The results of the user study show that variable-scale rate control can successfully improve an operators ability to control the position of the robot, without sacrificing the speed that the operator can complete a navigation task. However, as noted in Section \ref{sec:discussion}, a limitation of the developed method is that it is difficult for the operator to control the motion of the robot when the risk of a collision transitions from high to low, which rapidly increases the scale factor. In the future, we will improve our method by setting limits on the rate with which the scale factor can increase. 

In this paper, we investigate the operator's ability to control the robot through a two-dimensional environment, which they viewed from a bird's-eye perspective. In the future, we will test the developed methods in a three-dimensional simulated environment and will provide the operators with a first-person view via a simulated camera on the robot. In the future, we will also test the developed methods using a real UAV. The developed variable-scale rate control methods will theoretically work using data from LIDAR sensors or point cloud data from a map of the robot's environment generated online. Error from real sensor measurements could make it possible for the human operator to collide the UAV with an object, if the object's position is not accurately measured by the sensor. This may make it necessary to enlarge the critical region to ensure collision-free teleoperation. On the other hand, variable-scale rate control could prevent the operator from flying the UAV to a location that is incorrectly seen as occupied by the on-board sensors. This may make it necessary to enable the human operator to override the variable-scale rate control. 

Finally, we note that the risk factor used in the variable-scaling rate-control methods in this paper is based on the risk factors previously used to generate haptic feedback to help inform an human operator about the location of objects in the robot's environment \cite{Lam2009}. If haptic feedback is implemented with variable-scaling rate control, then the haptic feedback could inform the user both about the state of the robot's environment and the magnitude of the scale-factor. Therefore, we believe that haptic feedback could enhance the utility of variable-scaling rate control for teleoperation.



\bibliographystyle{IEEEtran} 
\bibliography{paper}

\end{document}